\documentclass[journal ]{new-aiaa}
\usepackage[utf8]{inputenc}
\usepackage{textcomp,dsfont,float}

\usepackage{graphicx,subfigure}
\usepackage{amsmath}
\usepackage[version=4]{mhchem}
\usepackage{siunitx}
\usepackage{longtable,tabularx}
\usepackage{xcolor}

\title{Linear Quadratic Guidance Law for Joint Motion Planning of a Pursuer-Turret Assembly
\footnote{
The views expressed are those of the authors and do not reflect the official guidance or position of the United States Government, the Department of Defense  the United States Air Force, or the United States Space Force.

Distribution Statement A: Approved for public release.  Distribution is unlimited.  Case numbers AFRL-2023-4749 and 2023-0939.
}}

\author{Bhargav Jha \footnote{Postdoctoral fellow, Department of Electrical and Computer Engineering, Michigan State University, bhargav@msu.edu} and Shaunak D.~Bopardikar\footnote{Assistant Professor, Department of Electrical and Computer Engineering, Michigan State University, shaunak@msu.edu}}
\affil{Michigan State University, East Lansing, Michigan, United States - 48864}
\author{Alexander Von Moll\footnote{Aerospace Engineer, Control Science Center, 2210 8th St., alexander.von$\_$moll@afrl.af.mil}}
\author{David Casbeer\footnote{Senior Electronics Engineer, Control Science Center, 2210 8th St., david.casbeer@afrl.af.mil AIAA Associate Fellow}}
\affil{Air Force Research Laboratory, WPAFB, OH, 45433}

\begin{document}

\maketitle

\section{Introduction}
A ship with a firing turret, mobile air-defense system, surveillance aircraft, and military vehicles with directed sensors are all examples of moving pursuers with rotating platforms installed onboard. The rotating platforms can be a turret, a missile launcher, or a gimballed camera or sensor. Various aspects of such systems have been studied in works such as \cite{schonberger1982flow,craig1981study,gupta1999evaluation,dos2005gun}. Recently, such systems have become increasingly fast and autonomous, thereby driving advancements in guidance and control of such vehicles. 

Classical guidance laws such as pure-pursuit, proportional navigation (PN), and line-of-sight guidance implement an underlying geometrical rule that is guaranteed to lead to interception. The requirement of minimal sensor information such as the line-of-sight measurement along with their computational efficiency makes these guidance laws popular for implementation. Under the assumption of a target moving at constant speed and small lead angles, the engagement kinematics can be linearized about the collision triangle, and the PN guidance law with a gain $(N=3)$ is shown to be energy optimal in \cite{bryson1965linear}. Using the linear approximation, for other gains, PN was shown to be optimal for slightly modified state-dependent cost in \cite{kreindler1973optimality}. The validity of the linearization for high-speed pursuers in planar engagements has also motivated the development of optimal guidance laws for scenarios such as interception of maneuvering targets \cite{nesline1981new}, intercept angle guidance \cite{shaferman2008linear}, pursuit-evasion differential games \cite{gutman1979optimal}, obstacle avoidance \cite{weiss2018linear}, aerial refuelling \cite{tsukerman2018optimal}, and cooperative guidance laws \cite{tan2018cooperative,jha2019cooperative}.

The above guidance laws aim for point capture of the target with near zero miss-distance. This is often conservative in a real world scenario. For instance, in a surveillance mission, it is sufficient to be in a finite region around the target point to survey it. In interception, the warheads used for neutralizing targets have a finite size and the interception is considered successful if it detonates in this region. Addressing this issue, works such as \cite{obermeyer2009path,dumitrescu2003approximation,obermeyer2010sampling,isaacs2011algorithms,weiss2016minimum} define capture as an inequality constraint of the miss distance from the target being less than a specified quantity. Doing this leads to improved optimality of the cost function \cite{weiss2016minimum}, which may be expressed in terms of the control effort of the pursuer or the final time of capture.
In \cite{obermeyer2009path,obermeyer2010sampling,isaacs2011algorithms}, the objective is to be in a neighborhood around a sequence of target points in minimum time. The pursuer is assumed to be a non-holonomic vehicle with saturated control and the targets are considered to be stationary. For a constantly maneuvering target and the linearized enagement model, \cite{weiss2016minimum} proposes analytic expressions for a closed loop state feedback guidance law with a specified miss distance. The low computational expense of the  guidance laws having analytic state-feedback form allows for their real time onboard implementation. 

Some applications also demand that the pursuer approaches the target point at a specified angle \cite{shaferman2008linear,dubins1957curves,ratnoo2008impact,gopalan2017generalized,ratnoo2009state}. This can be applicable to scenarios such as imaging a target from a certain orientation, exploiting the vulnerability of a target, or interception of a target while preventing collateral damage. Akin to the aforementioned case of point capture, achieving a perfect intercept angle may also be conservative. This is due to the fact that often a vulnerable region of a target is specified as a finite region, or a camera has a finite field of view instead of a single ray pointing to the target. This creates a scope for achieving a more optimal trajectory as compared to the perfect intercept angle constraint. Catering to this works such as \cite{indig2016near,vana2018optimal,manyam2017tightly} provide significant insight into these problems. In \cite{manyam2017tightly,vana2018optimal}, for a planar engagement, the time-optimal trajectory through a sequence of points and specified terminal cones for a constant speed non-holonomic pursuer with bounded controls  are proposed. For a three-dimensional case, \cite{indig2016near} proposes a heuristic approximation of time-optimal trajectory to a spatial point with an associated terminal cone. 

From the arguments in the aforementioned works, it is evident that instead of a point constraint, an inequality constraint on the terminal miss distance or the angle at which the pursuer approaches the target is more realistic. In this work, we consider a pursuer with an attached rotating platform to it. The axis of rotation is perpendicular to the plane of the vehicle. The rotating platform is a turret-like heavy machinery that needs to rotate and orient itself such that a constant maneuvering target, \textit{i.e.} a target with a constant lateral acceleration,  lies in the terminal cone associated with the turret's field-of-view. Moreover, in order to capture the target, the pursuer-turret assembly
has to reach within a user-specified range from the target. The specified range may signify the firing range of the turret, the effective range of a direct energy sensor, or the range of camera equipment installed on a surveillance aircraft. The slew dynamics of the turret as well as the lateral acceleration dynamics of the pursuer are assumed to be an arbitrary order linear time invariant dynamics. 

For high-speed pursuers with a heavy turret assembly, the change in the orientation of the turret due to the slewing of the turret becomes comparable to the change in its orientation due to the turn of the pursuer. This creates an opportunity for cooperation between the pursuer and the turret, $\textit{i.e.}$, depending on their rotational abilities, the turret, and the pursuer share the responsibility for the change in the turret's orientation. To facilitate this, we propose a minimum effort guidance law to capture a constant maneuvering target. Instead of the usual point constraints, the capture is defined as terminal state inequality constraints on the turret's orientation as well as miss distance from the target. The rest of the note is organized as follows. In Section \ref{section1}, we present the mathematical model of the pursuer-turret assembly and describe the non-linear engagement geometry. Thereafter, we present a linearization of the engagement geometry around the collision triangle. Using this linearization, in Section \ref{section2}, we formulate the optimal control problem and obtain an analytical expression of the guidance law. Although the analytical expression can be obtained for any arbitrary order linear dynamics of the pursuer-turret assembly, we exemplify a particular case where both the lateral acceleration dynamics of the pursuer and the turret's rotation dynamics are assumed to be ideal dynamics. To validate the applicability of the proposed guidance law, we present numerical simulations in Section \ref{section3} showing the cooperative behavior of the pursuer-turret assembly along with the effect of the parameters such as target speed, acceleration, and initial launch angle. 

\section{Engagement Geometry}\label{section1}

\subsection{Non-linear Engagement Geometry}
\begin{figure}
    \centering    \includegraphics[width=0.5\textwidth]{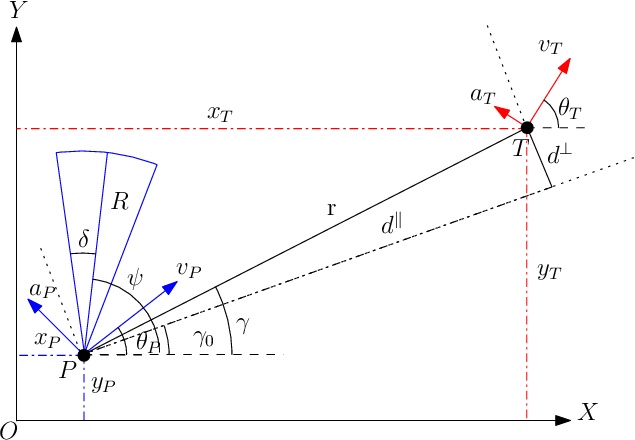}
    \caption{Engagement Geometry}
    \label{engagement_geometry}
\end{figure}

Consider a Cartesian coordinate system $XOY$. As shown in Fig. \ref{engagement_geometry}, we assume a planar engagement scenario consisting of two point objects namely a pursuer vehicle $(P)$ and a target vehicle $(T)$. The speeds of the pursuer and the target are assumed to be constant and denoted as $v_P$ and $v_T$, respectively. $\theta_P$ and $\theta_T$ are the heading angles of $P$ and $T$, respectively measured from positive X-axis in anti-clockwise direction. Both the vehicles can apply lateral accelerations $a_P$ and $a_T$ to change their respective heading angles.  The pursuer is also equipped with a forward-facing rotating turret $(C)$. Because the turret is assumed to be centered on the pursuer, their positions are identically given by $(x_P,y_P)\in \mathds{R}^2$. The sensing region of the turret is constrained by its maximum field-of-view (FOV) angle $\delta$ and the maximum range of detection $R$. As shown in Fig. \ref{engagement_geometry}, these two constraints lead to a horizontal conical sensing region of the turret. The angle of the line bisecting the FOV angle measured from the positive $X$-axis is denoted by $\psi$. An inertia-normalized torque $\tau$ can be applied to slew the turret at an angular speed $\omega$.

The kinematics of $P$ equipped with the turret is given by the following set of ordinary differential equations,
\begin{subequations}\label{pursuer}
\begin{align}
    \dot{x}_P &= v_P \cos\theta_P \\
    \dot{y}_P &= v_P \sin\theta_P \\
    \dot{\theta}_P &= \frac{{a}_P}{v_P} \\
        \dot{\psi} &= \frac{{a}_P}{v_P}+\omega \\
        \dot{\omega} &= \tau.
\end{align}
\end{subequations}
In any practical scenario, the lateral acceleration of $P$ will have an associated dynamics that maps it to the actuator command. Similarly, the rotating turret may also have some dynamics associated with its slewing. We assume these dynamics can be represented by arbitrary order linear time-invariant systems with realizations as,
\begin{align*}
    &u_1 \rightarrow a_P: \begin{array}{c|c} \mathbf{A_P} & \mathbf{B_P} \\ \hline \mathbf{C_P} & d_P \end{array}, 
    &u_2 \rightarrow \tau: \begin{array}{c|c} \mathbf{A_C} & \mathbf{B_C} \\ \hline \mathbf{C_C} & d_C \end{array}
\end{align*} 
where $u_1$ and $u_2$ are the control inputs to the pursuer and the turret.
Hence,
\begin{subequations}\label{dynamics}
\begin{align}
    \dot{\mathbf{x}}_P &= \mathbf{A_P} \mathbf{x}_P + \mathbf{B_P} u_1 \label{x1}\\
    a_P &= \mathbf{C_P} \mathbf{x}_P + d_P u_1 \label{x2} \\
    \dot{\mathbf{x}}_C &= \mathbf{A_C} \mathbf{x}_C + \mathbf{B_C} u_2 \label{x3}\\
    \tau &= \mathbf{C_C} \mathbf{x}_C + d_C u_2. \label{x4}
\end{align}
\end{subequations}
Here, $\mathbf{x}_P$ and $\mathbf{x}_C$ are states pertaining to dynamics of the pursuer's and turret's command, respectively. 

The kinematics of $T$ is given by,
\begin{align}\label{target}
    \dot{x}_T = v_T \cos\theta_T,~~~~~
    \dot{y}_T = v_T\sin\theta_T, ~~~~~\dot{\theta}_T = \frac{a_T}{v_T}.
\end{align}

We assume that the target acceleration throughout the engagement is constant and known \textit{a priori}. 
Let us now define a relative polar coordinate system $(r,\gamma)$ centered at $P$ such that,
\begin{align*}
x_T - x_P = r\cos\gamma,~~~~~~~~~~y_T - y_P = r\sin\gamma.
\end{align*}
Here, $r\geq 0$ is the line-of-sight (LOS) range and $\gamma\in[-\pi,\pi]$ is the LOS angle. Representing the engagement kinematics in this coordinate system, we have
\begin{subequations}\label{nonlinear}
\begin{align}
\dot{r} &= v_T \cos(\theta_T-\gamma) - v_P \cos(\theta_P-\gamma) \\
\dot{\gamma} &= \frac{1}{r}\left(v_T \sin(\theta_T-\gamma) - v_P \sin(\theta_P-\gamma)\right) \\
\dot{\theta}_P &= \frac{a_P}{v_P} \\
\dot{\psi} &= \frac{a_P}{v_P} + \omega \\
\dot{\omega} &= \tau \\
\dot{\theta}_T &= \frac{a_T}{v_T}.
\end{align}
\end{subequations}
In the proposed setup, there are two control inputs $u_1$ and $u_2$ (Eq.~\eqref{dynamics}) and the objective is to bring a maneuvering target in the sensing region of the turret. It is desired that during an engagement, $|u_1|$ and $|u_2|$ remain within their respective bounds of $\bar{u}_1$ and $\bar{u}_2$. Therefore, we devise a minimum-effort cooperative strategy such that the turret and the pursuer steer simultaneously, while compensating for their respective maneuverability and bring the target vehicle in the turret's sensing range and field of view. 
A suitable cost function to be minimized is,
\begin{align}\label{cost}
    J(u_1,u_2) \triangleq \frac{1}{2}\int^{t_f}_{t_0} \alpha \left(\frac{u_1}{\bar{u}_1}\right)^2 + (1-\alpha)\left(\frac{u_2}{\bar{u}_2}\right)^2 \mathrm{d}t
\end{align}
where $\alpha \in (0,1)$ and $t_f$ denotes the final time of capture. Here,  $\alpha\rightarrow 0$ depicts the case when the turret has sufficient ability to orient itself and $\alpha\rightarrow 1$ depicts the case when the turret's ability to orient itself is limited. The later case is akin to a turret being fixed rigidly on a moving pursuer. 

\subsection{Linearized Engagement Geometry}
The proposed optimal control problem is non-linear due to the engagement kinematics and does not admit a closed form feedback expression. Although non-linear optimal control problems can be solved using numerical methods, such solutions are difficult to implement onboard, and their analysis is quite tedious. Therefore, we resort to the linearization of the non-linear engagement kinematics around the initial collision triangle. We assume that the vehicles do not deviate significantly from the initial collision course and the corresponding lead angles $(\theta_i-\gamma)$ for $i\in\{P,T\}$ are small.  Such assumptions hold well for high-speed vehicles pursuing a relatively far away target and in scenarios where PN-like guidance laws are implemented \cite{kreindler1973optimality,nesline1981new,shaferman2008linear,gutman1979optimal}. Moreover, in the PN guidance law, the underlying geometrical rule is parallel navigation where the LOS is parallel to the initial LOS, and hence it does not rotate throughout the engagement.  The applicability of such an assumption in scenarios when the LOS rotates will be verified later in the paper. 

Let us denote the initial LOS angle as $\gamma_0$ and initial LOS range as $r_0$. As shown in Fig. \ref{engagement_geometry}, the relative separation parallel and perpendicular to the initial LOS are denoted as $d^\parallel$ and $d^\perp$, respectively. Hence, we have
\begin{align}\label{eomLin1}
    \ddot{d}^\perp = a_T \cos\left(\theta_{T0} - \gamma_0\right)-a_P \cos\left(\theta_{P0} - \gamma_0\right).
\end{align}
Note that as the lead angles are small, $d^\parallel$ is not significantly affected by the relative acceleration of the vehicles. Let us define a terminal time $t_f$ as the time when $d^\parallel(t)=0$. As $d^\parallel(t_0)=r_0$, we have
\begin{align}\label{eomLin2}
    t_f \triangleq \frac{r_0}{v_c},
\end{align}
where $v_c$ is the closing speed given by,
\begin{align}\label{eomLin3}
    v_c &= v_P \cos(\theta_{P0}-\gamma_0) - v_T \cos(\theta_{T0}-\gamma_0).
\end{align}
In order to approach the target such that the range and FOV limits of the turret are satisfied, the following terminal constraints need to be satisfied.
For the sensing range constraint,
\begin{align}\label{range_nonZEM}
    |d^\perp(t_f)| \leq R. 
\end{align}
At the terminal time, the constraint of the target being in the turret's field-of-view can be formulated as,
\begin{align}\label{FOV1_nonZEM}
    &\psi(t_f) - \delta \leq \gamma(t_f) \leq \psi(t_f) + \delta.
\end{align}
During the later stages of any guidance engagement to intercept a non-maneuvering target, a PN guidance law will lead to a non-rotating LOS \cite{kreindler1973optimality}, thereby validating the linearization near initial collision triangle. A similar linearization has also been found to be effective for the interception of a maneuvering target \cite{nesline1981new,garber1968optimum}, interception of maneuvering targets with a specified impact angle \cite{shaferman2008linear}, and even in engagements with multiple entities \cite{rubinsky2014three,tan2018cooperative}. The effectiveness of such a linearization is mainly due to the fact that in most of the above works, authors propose analytical expressions for the guidance laws which inherently facilitate recursive linearization of the engagement geometry at each time instant. In addition to this, for fast moving pursuers if one obtains a PN-like guidance law, the LOS does not rotate significantly. Therefore, we can assume that
\begin{align*}
    \gamma(t_f) \approx \gamma(t_0).
\end{align*}
Now the FOV constraint (Eq. \eqref{FOV1_nonZEM}  can be rewritten as 
\begin{align} \label{FOV2}
-\delta \leq \gamma(t_0) - \psi(t_f) \leq \delta.
\end{align}
\section{Optimal Control Problem}\label{section2}
Let $\mathbf{x} = \begin{bmatrix} d^\perp & \dot{d}^\perp & \psi & \omega &\bf{x}^\top_P & \bf{x}^\top_C \end{bmatrix}^\top\in \mathds{R}^{4+n_p+n_c}$ be the state vector and $\mathbf{u}=\begin{bmatrix}
    u_1 & u_2
\end{bmatrix}^\top\in\mathds{R}^2$ be the control input vector. The linearized equations of motion (EOM) described in Eqs. \eqref{dynamics} and Eqs. \eqref{eomLin1}-\eqref{eomLin3} can be rewritten as 
\begin{align}\label{linearKinematics}
    \dot{\mathbf{x}} = \mathbf{A} \mathbf{x} + \mathbf{B_1} \mathbf{u} + \mathbf{B_2} a_T.
\end{align}
where,
\begin{align*}
    &\mathbf{A}= \begin{bmatrix}
        0 & 1 & 0 & 0 & \mathbf{0} & \mathbf{0} \\
        0 & 0 & 0 & 0& -\cos(\theta_{P0}-\gamma_0) \mathbf{C_P} & \mathbf{0} \\
        0 & 0 & 0 & 1& \frac{1}{v_p}\mathbf{C_P} & \mathbf{0} \\
        0 & 0 & 0 & 0& \mathbf{0} & \mathbf{C_C}\\
        \mathbf{0} & \mathbf{0} & \mathbf{0} & \mathbf{0}& \mathbf{A_P} & \mathbf{0} \\
        \mathbf{0} & \mathbf{0} & \mathbf{0} & \mathbf{0} & \mathbf{0} & \mathbf{A_C}
    \end{bmatrix},~~
    &\mathbf{B_1}= \begin{bmatrix}
        0 & 0 \\
        -\cos\left(\theta_{P0}-\gamma_0\right) d_P & 0 \\
        \frac{1}{v_P}d_P & 0 \\
        0 & d_C \\
        \mathbf{B_P} & \mathbf{0} \\
        \mathbf{0} & \mathbf{B_C}
    \end{bmatrix},~~\mathbf{B_2}= \begin{bmatrix}
        0 \\ \cos(\theta_{T0} - \gamma_0) \\ 0 \\ 0\\ \mathbf{0} \\ \mathbf{0}
    \end{bmatrix}.
\end{align*}

\subsection{Optimal Guidance Law Derivation}
The dimension of the state vector $\mathbf{x}$ for the linearized EOM is $n_p+n_c+4$ and the constraints expressed in Eq. \eqref{range_nonZEM} and Eq. \eqref{FOV2} require only $2$ state variables. By using the terminal projection transformation \cite[Chapter 5]{bryson2018applied}, we can reduce the order of the system from $n_p+n_c+4$ to $2$.

 If the target is constantly maneuvering, \textit{i.e.}, $a_T$ is constant, for all times $t\in[t_0,t_f]$, consider the prediction of the complete state-vector at the final time. Then 
\begin{equation}\label{linearModelRealCoordinates}
    \mathbf{x}(t_f,t) = \Phi(t_f,t)\mathbf{x}(t) + \int^{t_f}_{t} \Phi(t_f, \tau) \mathbf{B_2} a_T ~\mathrm{d}\tau,
\end{equation}
where $\Phi(t_f,t)$ is the state transition matrix associated with matrix $\mathbf{A}$ and is given by,

\begin{align*}
    \Phi(t_f,t) \triangleq \exp{\mathbf{A}(t_f-t)} = \begin{bmatrix}
        1 & t_f-t & 0 & 0 & \boldsymbol{\phi}_{15} & \boldsymbol{0} \\
        0 & 1 & 0 & 0 &  \boldsymbol{\phi}_{25} & \boldsymbol{0} \\
        0 & 0 & 1 & t_f-t & \boldsymbol{\phi}_{35} & \boldsymbol{\phi}_{36} \\
        0 & 0 & 0 & 1 & \boldsymbol{0} & \boldsymbol{\phi}_{46} \\
        \boldsymbol{0} & \boldsymbol{0} & \boldsymbol{0} & \boldsymbol{0} &\boldsymbol{\phi}_{55} & \boldsymbol{0} \\
        \boldsymbol{0} & \boldsymbol{0} & \boldsymbol{0} & \boldsymbol{0}& \boldsymbol{0} & \boldsymbol{\phi}_{66}
    \end{bmatrix}.
\end{align*}
The elements $\boldsymbol{\phi}_{ij}$, for $i,j\in \mathds{Z}_{[1,~4+n_p+n_c]}$ are not specified as they depend on the particular dynamics of the pursuer-turret assembly as described in Eq. \eqref{dynamics}.

As the terminal constraints are specified only for $d^\perp$ and $\gamma_0-\psi$, we are interested in the prediction of these quantities at the final time. Now, for any time $t \leq t_f$, denote the prediction of $d^\perp$ and $\gamma_0-\psi$ at final time as $z_d(t)$ and $z_\psi(t)$, respectively. These can be obtained from the first two components of $\mathbf{x}(t_f,t)$ as,
\begin{align}
    z_d(t)&= \mathbf{D_P}\left(\Phi(t_f,t)\mathbf{x}(t) + \int^{t_f}_{t} \Phi(t_f, \tau) \mathbf{B_2} a_T ~\mathrm{d}\tau\right) \label{zdMat}\\
    z_\psi(t)&=\gamma_0-\mathbf{D_C}\left(\Phi(t_f,t)\mathbf{x}(t) + \int^{t_f}_{t} \Phi(t_f, \tau) \mathbf{B_2} a_T ~\mathrm{d}\tau\right) \label{zPsiMat}
\end{align}
where $\mathbf{D_P}=\begin{bmatrix}
    1 & 0 & 0 & 0 & \mathbf{0} & \mathbf{0}

\end{bmatrix}$ and $\mathbf{D_C}=\begin{bmatrix}
    0 & 0 & 1 & 0 &\mathbf{0} & \mathbf{0}
\end{bmatrix}$.
This translates to
\begin{align}
 z_d(t) &= d^\perp(t) + \dot{d}^\perp(t) (t_f-t) + \boldsymbol{\phi}_{15}(t)\mathbf{x}_P + a_T\cos(\theta_{T0}-\gamma_0)\frac{(t_f-t)^2}{2} \label{zdEq}\\
    z_\psi(t) &= \gamma_0-\psi(t) - \omega(t)(t_f-t) - \boldsymbol{\phi}_{35}(t)\mathbf{x}_P - \boldsymbol{\phi}_{36}(t)\mathbf{x}_C \label{zPsiEq}.
\end{align}
Note that $z_d(t)$ is the zero-effort-miss distance, which is defined as the predicted separation between the pursuer and the target if the target maneuvers as expected and the pursuer-turret assembly does not apply any control from current time onward ($\mathbf{u}(t) \equiv \mathbf{0} ~ \forall t\in[t,t_f]$). Under the same conditions, $z_\psi(t)$ is the predicted turret angle at the terminal time.  
Using the fact that
\begin{align*}
    \dot{\Phi}(t_f,t) &= -\Phi(t_f,t)\mathbf{A}, &\Phi(t_f,t_f) = \mathbf{I} 
\end{align*}
we can compute the time-derivatives of the predicted states $z_d(t)$ and $z_\psi(t)$ from Eqs. \eqref{zdMat}-\eqref{zPsiMat} and obtain,
\begin{align*}
    \dot{z}_d(t) &= \left(-d_P(t_f-t)\cos(\theta_{P0}-\gamma_0) +\boldsymbol{\phi}_{15}{\mathbf{B_P}}\right)u_1 \\
    \dot{z}_\psi(t) &= -\left(\boldsymbol{\phi}_{35}\mathbf{B_P} + \frac{d_P}{v_P}\right)u_1- \left(\boldsymbol{\phi}_{36}\mathbf{B_C} + (t_f-t)d_C\right)u_2.
\end{align*}
We now redefine the optimal control problem in terms of the transformed states. Let $\mathbf{z} = \begin{bmatrix}
    z_d & z_\psi
\end{bmatrix}^\top$. The transformed state dynamics are now defined as,
\begin{align}
    \dot{\mathbf{z}} = \mathbf{B}_z(t) \mathbf{u} \label{zState}
\end{align}
where,
$$\mathbf{B}_z(t)=\begin{bmatrix}
    \left(-d_P(t_f-t)\cos(\theta_{P0}-\gamma_0) +\boldsymbol{\phi}_{15}{\mathbf{B_P}}\right) & 0\\
    -\boldsymbol{\phi}_{35}\mathbf{B_P} - \frac{d_P}{v_P} & -\boldsymbol{\phi}_{36}\mathbf{B_C} - (t_f-t)d_C
\end{bmatrix}.$$
The transformed terminal  constraints can be expressed as,
\begin{subequations}
\begin{align}
    &|z_d(t_f)| \leq R \\
    &|z_\psi(t_f)| \leq \delta.
\end{align}\label{constraints}
\end{subequations}
The cost function remains unchanged as,
    \begin{align*}
    J(u_1,u_2) \triangleq \frac{1}{2}\int^{t_f}_{t_0} \alpha \left(\frac{u_1}{\bar{u}_1}\right)^2 + (1-\alpha)\left(\frac{u_2}{\bar{u}_2}\right)^2 \mathrm{d}t
\end{align*}

\subsection{Necessary Conditions for Optimality}
For the above optimal control problem, the Hamiltonian is defined as,
\begin{align*}
    H = \frac{1}{2}\mathbf{u}^\top \Sigma \mathbf{u} + \mathbf{p}^\top \mathbf{B}_z \mathbf{u}
\end{align*}
where $\mathbf{p}\triangleq \begin{bmatrix}
    p_d & p_\psi
\end{bmatrix}$ is the costate vector, and $p_d$ and $p_\psi$ are the costates corresponding to $z_d$ and $z_\psi$, respectively. $\boldsymbol{\Sigma}$ denotes the weighting matrix defined as $$\boldsymbol{\Sigma}\triangleq \text{diag}\left(\frac{\alpha}{\bar{u}^2_1},~\frac{1-\alpha}{\bar{u}^2_2}\right). $$ By the Maximum Principle \cite{pontryagin1987mathematical},
\begin{align}
    \dot{\mathbf{p}} = - \frac{\partial H}{\partial \mathbf{z}} = 0 \label{pConst}\\
    \mathbf{u} = - \Sigma^{-1} \mathbf{B}^\top_z(t)\mathbf{p}.  \label{control}
\end{align}
At the terminal time, let the transformed states 
\begin{align}
    &z_d(t_f) = c_d, &-R\leq c_d\leq R \label{range} \\
&z_\psi(t_f) = c_\psi, & -\delta \leq c_\psi \leq \delta. \label{FOV1}
\end{align}
Hence we have,
\begin{equation}
    \mathbf{z}(t_f) = \begin{bmatrix}
    c_d &
    c_\psi 
\end{bmatrix}^\top
\end{equation}
Note that the costate vector $\mathbf{p}$ is constant due to Eq. \eqref{pConst}. By substituting the value of the control input from Eq. \eqref{control} into Eq. \eqref{zState} and integrating it we obtain,
\begin{equation}
    \mathbf{z}(t_f) - \mathbf{z}(t_0) = \int^{t_f}_{t_0} -\mathbf{B}_z(\tau)\Sigma^{-1}\mathbf{B}^\top_z(\tau)\mathbf{p}~\mathrm{d}\tau.
\end{equation}
This can be rewritten as
\begin{equation}\label{GMat}
    \mathbf{z}(t_f) = \mathbf{z}(t_0) + \mathbf{G} \mathbf{p}
\end{equation},
where 
$$\mathbf{G}= \int^{t_f}_{t_0} -\mathbf{B}_z(\tau)\Sigma^{-1}\mathbf{B}^\top_z(\tau)~\mathrm{d}\tau = \begin{bmatrix}
    G_{11} & G_{12} \\ G_{12} & G_{22}
\end{bmatrix}.$$
Here, $G_{ij}$ for $i,j\in\{1,2\}$ denote the components of an invertible negative definite matrix $\mathbf{G}$ which can be computed explicitly for any given dynamics expressed in the form of Eq. \eqref{x1}-\eqref{x4}.
Hence,
\begin{align}
    \mathbf{p} = \mathbf{G}^{-1}\left(\mathbf{z}(t_f) - \mathbf{z}(t_0)\right). \label{lambdaEq}
\end{align}
Substituting Eqs. \eqref{control}-\eqref{lambdaEq} into the cost function in Eq. \eqref{cost} 
\begin{align}
    J &= \frac{1}{2}\int^{t_f}_{t_0} \mathbf{p}^\top \mathbf{B}_z(t) \Sigma^{-1} \mathbf{B}^\top_z(t)  \mathbf{p}~ \mathrm{d}t \nonumber \\
    &= -\frac{1}{2} \mathbf{p}^\top \mathbf{G} \mathbf{p} \nonumber \\
    &= -\frac{1}{2} \left(\mathbf{G}^{-1}\left(\mathbf{z}(t_f) - \mathbf{z}(t_0)\right)\right)^\top \mathbf{G} \mathbf{G}^{-1}\left(\mathbf{z}(t_f) - \mathbf{z}(t_0)\right) \nonumber \\
    &=-\frac{1}{2}\left(\mathbf{z}(t_f) - \mathbf{z}(t_0)\right)^\top \mathbf{G}^{-1} \left(\mathbf{z}(t_f) - \mathbf{z}(t_0)\right) .\label{cost2}
\end{align}
Note that Eq. \eqref{cost2}, is a quadratic function of constants $c_d$ and $c_\psi$. As both these constants are bounded (see Eq. \eqref{range}-\eqref{FOV1}), the optimal value of these constants ($c^*_d$, $c^*_\psi$) can be computed by solving the following quadratic optimization problem,
\begin{align}\label{opt}
   c^*_d,\ c^*_\psi \in \underset{|c_d|\leq R,~|c_\psi|\leq \delta}{\arg \min} -\left(\begin{bmatrix}
        c_d \\ c_\psi
    \end{bmatrix} - \mathbf{z}(t_0)\right)^\top \mathbf{G}^{-1} \left(\begin{bmatrix}
        c_d \\ c_\psi
    \end{bmatrix} - \mathbf{z}(t_0)\right) 
\end{align}
where $\mathbf{G}^{-1}=\frac{1}{\Delta}\begin{bmatrix}
    {G}_{22} & -{G}_{12} \\ -{G}_{12} & {G}_{11}
\end{bmatrix}$
and $\Delta \triangleq {G}_{22}{G}_{11}-{G}^2_{12}$. As the cost function $J$ and the constraints are convex in the variables $c_d$ and $c_\psi$, the minima of the above quadratic optimization problem will exist provided $\mathbf{G} \prec 0$. Let $\lambda_1,~\lambda_2,~\lambda_3,~$and $\lambda_4$ be the Lagrange multipliers associated with the constraints $c_d-R\leq 0$, $-c_d-R\leq 0$, $c_\psi- \delta\leq 0$, and $-c_\psi- \delta\leq 0$, respectively. For this constrained problem, the augmented cost can be written as,
\begin{align*}
L = \frac{1}{\Delta}\left(-G_{22}(c_d-z_d(t_0))^2 -G_{11}(c_\psi-z_\psi(t_0))^2 + 2 G_{12}(c_d-z_d(t_0))(c_\psi-z_\psi(t_0))\right) + \nonumber \\ \lambda_1(c_d-R) + \lambda_2(-c_d-R) + \lambda_3(c_\psi-\delta)+ \lambda_4(-c_\psi-\delta).
\end{align*}
Note that the constraints $c_d-R\leq0$ and $-c_d-R\leq 0$ cannot be active simultaneously. Similarly, $c_\psi-\delta\leq0$ and $-c_\psi- \delta\leq 0$ also can not be active simultaneously. Now depending on which of the constraints is active or inactive and by the complementary slackness condition, the global minimizer will result from one of the following nine cases,
\begin{align*}
    &\left(\lambda_1 = \lambda_2 = \lambda_3 = \lambda_4 = 0\right),  & \left(c_d = z_d(t_0),\ c_\psi = z_\psi(t_0)\right),~~~\text{if } \left|z_d(t_0)\right|\leq R, \text{ and }\left|z_\psi(t_0)\right|\leq \delta \\
    &\left(\lambda_1 > 0,~ \lambda_2 = 0,~  \lambda_3 =0,~ \lambda_4 =0\right), &\left(c_d = R,\ c_\psi = c_\psi(R)\right),~~~\text{if } \left|c_\psi(R)\right|\leq \delta\\
    &\left(\lambda_1 > 0,~ \lambda_2 = 0,~  \lambda_3 >0,~ \lambda_4 = 0\right), &\left(c_d = R,\ c_\psi = \delta\right)\\
&\left(\lambda_1 > 0,~ \lambda_2 = 0,~  \lambda_3 =0,~ \lambda_4 > 0\right), &\left(c_d = R,\ c_\psi = -\delta\right)\\
&\left(\lambda_1 = 0,~ \lambda_2 > 0,~  \lambda_3 =0,~ \lambda_4 =0\right), &\left(c_d = -R,\ c_\psi = c_\psi(-R)\right),~~~\text{if } \left|c_\psi(-R)\right|\leq \delta\\
    &\left(\lambda_1 = 0,~ \lambda_2 > 0,~  \lambda_3 >0,~ \lambda_4 = 0\right), &\left(c_d = -R,\ c_\psi = \delta\right)\\
&\left(\lambda_1 = 0,~ \lambda_2 > 0,~  \lambda_3 =0,~ \lambda_4 > 0\right), &\left(c_d = -R,\ c_\psi = -\delta\right)\\
&    \left(\lambda_1 = 0,~ \lambda_2 = 0,~  \lambda_3 >0,~ \lambda_4 =0\right), & \left(c_d = c_d\left(\delta\right),\ c_\psi = \delta\right),~~~\text{if } \left|c_d\left(\delta\right)\right|\leq R\\
&    \left(\lambda_1 = 0,~ \lambda_2 = 0,~  \lambda_3 =0,~ \lambda_4 > 0\right) &\left(c_d = c_d\left(-\delta\right),\ c_\psi = -\delta\right),~~~\text{if } \left|c_d\left(-\delta\right)\right|\leq R.
\end{align*}
Here $c_d\left(\pm\delta\right)$ and $c_\psi\left(\pm R\right)$ can be computed from the solution of the first order necessary condition,
\begin{align*}
    \nabla L(\mathbf{c},\boldsymbol{\lambda}) = 0.
\end{align*}
$$c_\psi\left(\pm R\right) = z_\psi(t_0) + \frac{G_{12}}{G_{11}}\left(\pm R-z_d(t_0)\right),$$
$$c_d\left(\pm \delta\right) = z_d(t_0) + \frac{G_{12}}{G_{22}}\left(\pm \delta-z_\psi(t_0)\right).$$

The optimal value $\left(c^*_d,c^*_\psi\right)$ can be computed by finding the corresponding values of the objective function for each of the above nine cases. Substituting the minimum value leads to the closed form solution of the optimal guidance law,
\begin{align}\label{controls}
\mathbf{u}(t) = - \Sigma^{-1}\mathbf{B}^\top_z(t)\mathbf{G}^{-1}\left(\begin{bmatrix}
    c^*_d \\ c^*_\psi 
\end{bmatrix} - \begin{bmatrix}
    z_d(t_0) \\ z_\psi(t_0)
\end{bmatrix}\right)
\end{align}
The control inputs obtained above are for open-loop control for a given initial condition. These control inputs can be implemented in closed loop state-feedback if the values at the initial time are replaced by the values at current time. Hence, we can substitute the current time $t$ instead of $t_0$ and $t_{go}\triangleq t_f-t$ in Eq. \eqref{controls} to obtain
\begin{subequations}\label{controls1}
\begin{align}
    u_1\left(t_{go}, z_\psi(t), z_d(t)\right) &= \frac{\Delta \bar{u}^2_1}{\alpha}\left(\left(\Gamma_1 G_{12} - \Gamma_2 G_{11}\right) \left(c^*_\psi-z_\psi(t)\right) - \left(\Gamma_1 G_{22} - \Gamma_2 G_{12}\right) \left(c^*_d-z_d(t)\right)   \right) \\
    u_2\left(t_{go}, z_\psi(t), z_d(t)\right) &= \frac{\Delta \bar{u}^2_2}{1-\alpha}\left(-t_{go}d_C - \boldsymbol{\phi}_{36} \mathbf{B_C}\right)\left(G_{12}\left(c^*_d-z_d(t)\right)-G_{11}\left(c^*_\psi-z_\psi(t)\right)\right)
\end{align}
\end{subequations}
where $\Gamma_1 \triangleq -d_P t_{go} \cos(\theta_P-\gamma) + \boldsymbol{\phi}_{15}\mathbf{B_P}$ and $\Gamma_2 \triangleq -\boldsymbol{\phi}_{35}\mathbf{B}_P - \frac{d_P}{v_P}$.

\subsection{Implementation of the Guidance Law}
 The proposed guidance law require $z_d(t)$, $z_\psi(t)$, and $t_{go}$ as the measured states. As can be seen from Eq. \eqref{zdEq}-\eqref{zPsiEq}, these states are not directly obtained from the engagement kinematics. Hence, we need to approximate these states in terms of the states obtained from the engagement kinematics, \textit{i.e}, $r$ and $\lambda$. The other states such as $\mathbf{x_P}$ and $\mathbf{x_C}$ are related to the dynamics of the pursuer-turret assembly and are assumed to be available from onboard sensors.

The approximate time-to-go can be obtained as,
\begin{align}
    t_{go} = \frac{r(t)}{v_c(t)} \label{tgoApprox}
\end{align}
 Under the small angle approximations and previous assumptions for the linear engagement kinematics we have,
\begin{align*}
\gamma=\gamma_0 + \tan^{-1}\left(\frac{d^\perp}{d^\parallel}\right) \approx \gamma_0 + \frac{d^\perp}{d^\parallel}   
\end{align*}
Under this approximation, the time-derivative of LOS angle can be expressed as,
\begin{align*}
    \dot{\gamma} = \frac{\dot{d}^\perp d^\parallel -  d^\perp\dot{d}^\parallel}{(d^\parallel)^2} = \frac{\dot{d}^\perp v_c t_{go} + v_c d^\perp }{(v_c t_{go})^2} =  \frac{\dot{d}^\perp t_{go} +  d^\perp }{v_c t^2_{go}}
\end{align*}
Comparing with Eq. \eqref{zdEq}, the approximation of zero-effort-miss distance in terms of the measured state can be obtained as,
\begin{align*}
    z_d &\approx \dot{\gamma}v_c  + \frac{1}{2}a_Tt^2_{go} + \boldsymbol{\phi}_{15}(t_{go})\mathbf{x}_P. 
\end{align*}
As mentioned before, the transformed state $z_\psi(t)$ is the prediction of the difference between the LOS angle and $\psi$ assuming that the pursuer-turret system applies no control from current time onwards until the end of the engagement and the target maneuvers as expected. We obtain 
\begin{align*}
    \psi(t_f) &= \psi(t) + \omega t_{go} + \boldsymbol{\phi}_{35}(t_{go})\mathbf{x}_P + \boldsymbol{\phi}_{36}(t_{go})\mathbf{x}_C.
\end{align*}
In the derivation so far we have assumed that $\dot{\gamma} \equiv 0$, hence, one may obtain
    $z_\psi(t) = \gamma(t)-\psi(t_f)$.
This prediction is exact for engagements when both the pursuer-turret assembly and the target are on the collision course.  It may happen that the pursuer and target may not be exactly on the collision course causing the LOS to rotate and hence $\gamma(t_f) \not\approx \gamma(t_0)$. With a slight abuse of notation, let $\gamma(t,t_0):\mathds{R}^2\rightarrow \mathds{R}$ be a function denoting the evolution of the LOS angle with time, if no control was applied from  time $t_0$ onward.  The Taylor series expansion of $\gamma(t,t_0)$ about the time $t_0$ is given by
\begin{align*}
    \gamma(t,t_0) = \gamma(t_0) + \dot{\gamma}(t_0)(t-t_0) + \frac{\ddot{\gamma}(t_0)(t-t_0)^2}{2} + o(t)
\end{align*}
where $o(t)$ denotes the higher order terms.
Replacing $\gamma(t_f)$ with the first order approximation of $\gamma(t_f,t)$ in Eq. \eqref{zPsiEq} provides a better approximation of $z_\psi(t)$ as,
\begin{align*}
    z_\psi(t) &\approx \gamma(t) + \dot{\gamma}(t)t_{go} - \psi(t) - \omega t_{go} - \boldsymbol{\phi}_{35}(t_{go})\mathbf{x}_P - \boldsymbol{\phi}_{36}(t_{go})\mathbf{x}_C
\end{align*}
Note that the above prediction can be improved if the higher derivatives of $\gamma(t)$ are known, but these quantities are difficult to obtain from sensor measurements.

\subsection{Particular Case of Ideal Pursuer-Turret Dynamics}
Now we will present a particular case where both the pursuer's and the turret's rotation dynamics are ideal. Hence, we have the states and control inputs defined as,
\begin{align*}
    \mathbf{\bar{x}}&=\begin{bmatrix}
        d^\perp & \dot{d}^\perp & \psi & \omega
    \end{bmatrix}^\top &\mathbf{\bar{u}}=\begin{bmatrix}
        a_P & \tau
    \end{bmatrix}^\top
\end{align*}
The state dynamics is given by,
\begin{align*}
\dot{\bar{x}}_1 &= \dot{\bar{x}}_2 \\
\dot{\bar{x}}_2 &= a_T \cos(\theta_{T0}-\gamma_0) - a_P \cos(\theta_{P0}-\gamma_0) \\
\dot{\bar{x}}_3 &= \omega + \frac{a_P}{v_P} \\
\dot{\bar{x}}_4 &= \tau
\end{align*}
where $\bar{x}_i$ denote the $i^{th}$ component of $\mathbf{\bar{x}}$.

For this particular case, the cost function to minimize is expressed as $$J(a_p,\tau) = \int^{t_f}_0 a^2_P + \frac{(1-\alpha)\bar{u}^2_1}{\alpha \bar{u}^2_1} \tau^2 ~\mathrm{d}t.$$

%
Solving this problem in a similar way as before, from Eq. \eqref{GMat}, the matrix $\mathbf{G}$  is given as,
\begin{align*}
    &\mathbf{G} = \begin{bmatrix}
        -\frac{\bar{u}_1(t_f-t_0)^3}{3\alpha} \cos(\theta_{P0}-\gamma_0)^2 & -\frac{\bar{u}_1(t_f-t_0)^2}{2 \alpha v_P} \cos(\theta_{P0}-\gamma_0)  \\-\frac{\bar{u}_1(t_f-t_0)^2}{2 \alpha v_P} \cos(\theta_{P0}-\gamma_0) & -\frac{\bar{u}_1 (t_f-t_0)}{\alpha v^2_P} - \frac{\bar{u}_2(t_f-t_0)^3}{3(1-\alpha)}
      \end{bmatrix}
\end{align*}
Note that if $t_f>t_0$, $\mathbf{G} \prec 0$ and hence $\mathbf{G}$ invertible and the minima of the optimization problem in Eq. \eqref{opt} exists.
Now from Eq. \eqref{controls}, the control inputs can be obtained as
\begin{subequations}\label{controls3}
\begin{align}
    a_P &= -\frac{(18\sigma + 12 t^2_{go}v^2_P)(c^*_d-z_d)+6 \sigma \cos(\theta_{P}-\gamma) v_P t_{go} (c^*_\psi-z_\psi)}{3\sigma\cos(\theta_{P}-\gamma)t^2_{go} + 4  v^2_P \cos(\theta_{P}-\gamma)t^4_{go}}  \\
    \tau&=-\frac{-18  v_P (c^*_d-z_d) + 12  \cos(\theta_{P}-\gamma) t_{go} v^2_P (c^*_\psi-z_\psi)}{3\sigma\cos(\theta_{P}-\gamma)t_{go} + 4  v^2_P \cos(\theta_{P}-\gamma)t^3_{go}},
\end{align}
\end{subequations}
where $\sigma\triangleq\frac{(1-\alpha)\bar{u}^2_1}{\alpha \bar{u}^2_1}$.

Remark 1:  Note that as $\alpha\rightarrow 0$ ($\sigma\rightarrow \infty$), then the control effort associated with $\tau$ becomes increasingly expensive. Hence, from Eq. \eqref{controls3} $a_P \rightarrow \frac{6}{\cos(\theta_P-\gamma)t^2_{go}}\left(c^*_d-z_d\right) + \frac{2v_P}{t_{go}}\left(c^*_\psi-z_\psi\right)$ and $\tau\rightarrow 0$. This shows that the trajectory of the pursuer is altered to orient the turret with a limited rotational ability. 

Remark 2: As $\alpha \rightarrow 1$ ($\sigma\rightarrow 0$), the turret is fast enough to orient itself. Hence, for this case $a_P \rightarrow \frac{3}{\cos(\theta_P-\gamma)t^2_{go}}\left(c^*_d-z_d\right)$ and $\tau\rightarrow -\frac{-18  (c^*_d-z_d) + 12  \cos(\theta_{P}-\gamma) t_{go} v_P (c^*_\psi-z_\psi)}{4  v_P \cos(\theta_{P}-\gamma)t^3_{go}} $. Note that if $R\rightarrow 0$, then for this case $a_P \rightarrow -3\frac{z_d}{\cos(\theta_P-\gamma)t^2_{go}}$. This is the expression for the optimal control input for the augmented proportional navigation (APN) guidance law \cite{nesline1981new}. This shows that if the turret has a better rotational ability, it can orient itself such that the pursuer does not have to alter its trajectory to orient the turret, and hence saving the control effort associated with $a_P$. 

Hence, Remark 1 and Remark 2  highlight the benefits of the joint motion planning for the pursuer-turret assembly.

\section{Numerical Simulations}\label{section3}
To validate the performance of the guidance law and to study the effect of parameters, we present some numerical simulations. The guidance law obtained in Eq. \eqref{controls1} is obtained using the linearized engagement kinematics.
\begin{table}[!htbp]
    \centering
    \begin{tabular}{|c|c|}
    \hline
       Parameter  & Value \\
       \hline
          $(x_P(t_0),y_P(t_0),\theta_P(t_0))$      & $(0 ~ \rm{m},0~ \rm{m}, \frac{\pi}{4}~ \rm{rad})$ \\ \hline
          $(x_T(t_0),y_T(t_0),\theta_T(t_0))$      & $(5000 ~\rm{m},0~\rm{m},\frac{3\pi}{4}~ \rm{rad})$ \\ \hline
          $v_P$      & $400~ \rm{m~s}^{-1}$ \\ \hline
          $v_T$      & $350~ \rm{m~s}^{-1}$ \\ \hline
          $\psi(t_0)$ & $\frac{\pi}{2}~ \rm{rad}$ \\ \hline
          $\omega(t_0)$ & $0~ \rm{rad}~s^{-1}$ \\ \hline
          $\alpha$      & $0.5$ \\ \hline
          $R$      & $500 ~\rm{m}$ \\ \hline
          $\delta$      & $\frac{\pi}{4} ~\rm{rad}$ \\ \hline
          $\bar{u}_1$      & $1 $ \\ \hline
          $\bar{u}_2$      & $10^{-4} $ \\ \hline
    \end{tabular}
    \caption{Parameters}
    \label{parameterTable}
\end{table}

\begin{figure}[!htbp]
    \centering
\subfigure[Trajectory]{\includegraphics[width=0.35\textwidth]{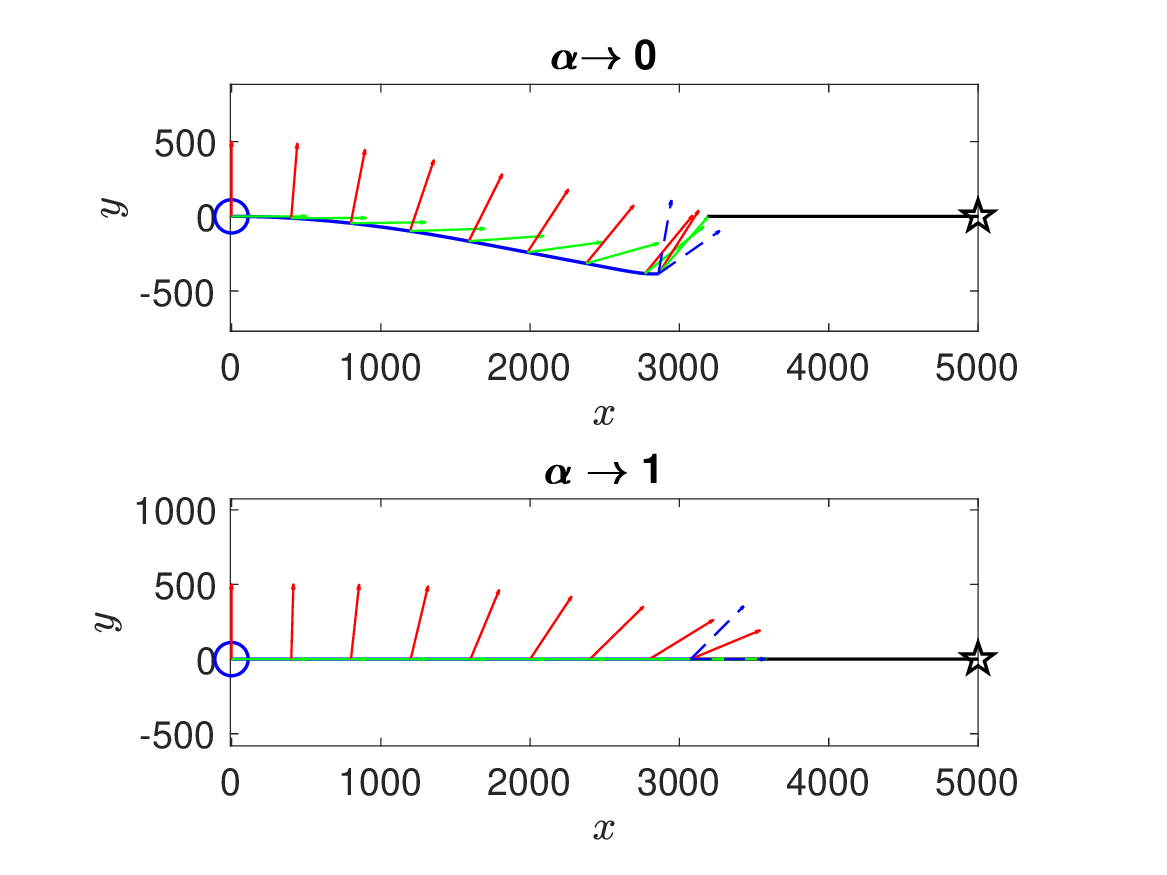}\label{variationsAlpha:Trajectory}}
\subfigure[Pursuer's acceleration]{\includegraphics[width=0.35\textwidth]{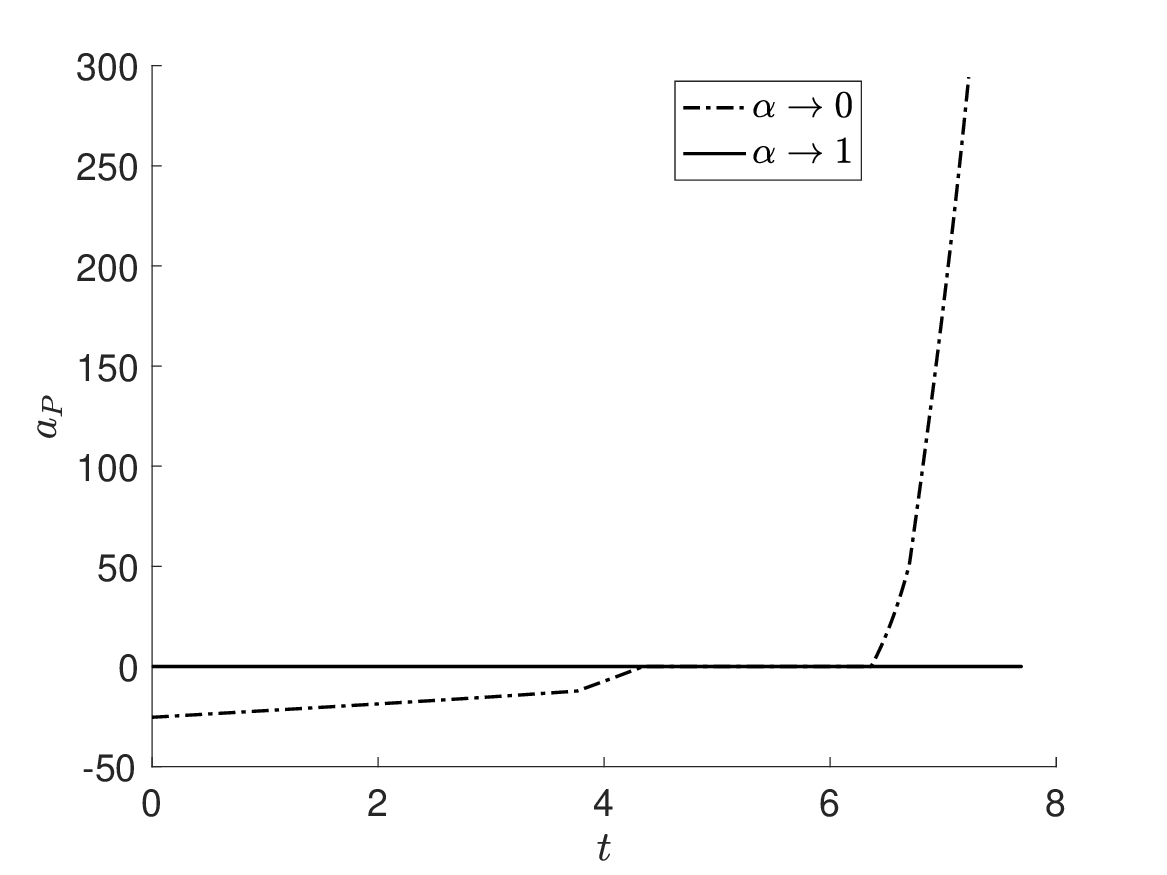}\label{variationsAlpha:latax}}
\subfigure[Turret's torque]{\includegraphics[width=0.35\textwidth]{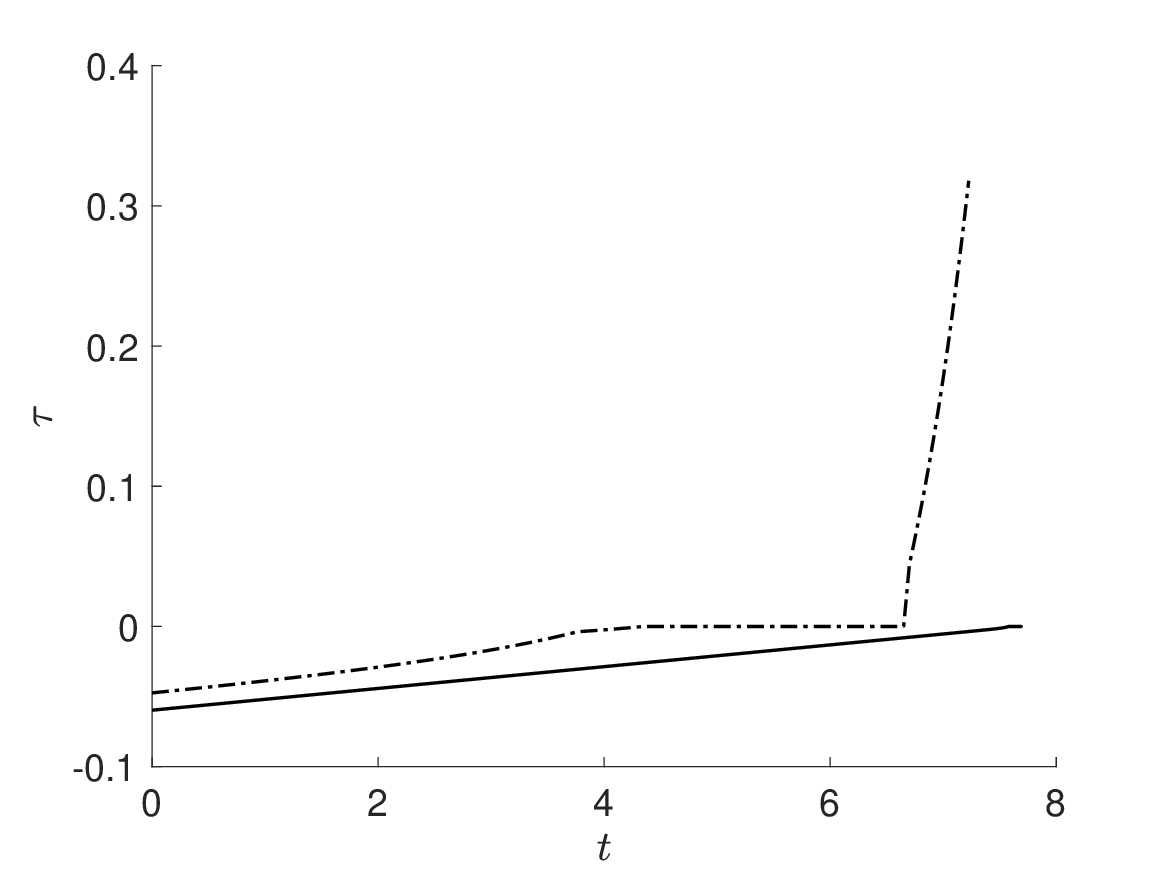}\label{variationsAlpha:torque}}
\subfigure[Normalized error]{\includegraphics[width=0.35\textwidth]{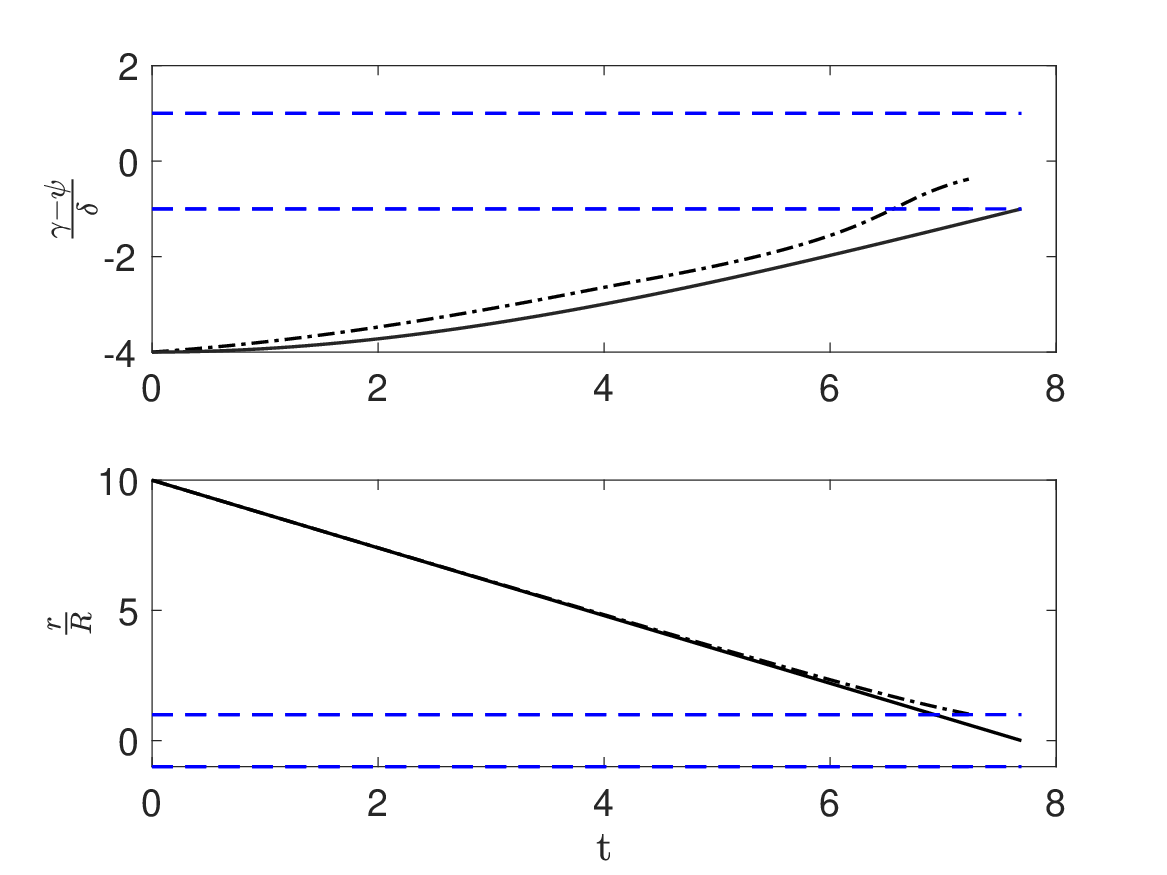}\label{variationsAlpha:error}}
    \caption{Variation of Trajectories with $\alpha$}
    \label{variationsAlpha}
\end{figure}
Hence, using them with the kinematics described in  Eq. \eqref{linearKinematics} will lead to capture at the prescribed terminal time $t_f$. 

But it is required to verify the applicability of the linear guidance law when the non-linear engagement kinematics given in Eqs. \eqref{pursuer}-\eqref{target} is considered. Hence, we demonstrate the performance of the guidance law considering the nonlinear engagement kinematics  along with ideal dynamics for $a_P$ and $\tau$ are used. Unless stated otherwise, the values and the units of the parameters used in the simulations are as described in Table. \ref{parameterTable}.
In all the $XY$ trajectory plots, the pursuer and evader trajectories are shown by blue line and black line, respectively. The initial positions of the pursuer and target are marked by a circle and a pentagram, respectively. Throughout the pursuer's trajectory, vectors along the instantaneous LOS are denoted by green lines and the vectors of length $R$ along the instantaneous $\psi$ are denoted by red lines. At the time of capture, the FOV and the sensing range of the pursuer-turret assembly is represented by two dashed blue lines of length $R$. 

Fig. \ref{variationsAlpha} exemplifies the advantage of joint motion planning of pursuer-turret assembly. In Fig. \ref{variationsAlpha:Trajectory}, for the case of $\alpha\rightarrow 1$, the turret rotates throughout the engagement and the capture is achieved without the pursuer's maneuver. This shows the role of the turret in minimizing the lateral acceleration effort of the pursuer. If $\alpha\rightarrow 0$, the pursuer-turret assembly prioritizes reduction of the control effort of the camera and hence the pursuer maneuvers to orient the turret for the final capture. Note that for $\alpha\rightarrow 0$, the zero-effort-miss $z_d$ was $0$ at the beginning of the engagement while the minimum required $z_d$ is $R$. The pursuer uses this leverage to compensate for the limited ability of the turret to orient itself. For both the cases, Fig. \ref{variationsAlpha:latax} and Fig. \ref{variationsAlpha:torque} shows the lateral acceleration and torque profile, respectively. The successful capture can also be verified by the normalized range $\frac{r}{R}$ and normalized orientation $\frac{\gamma-\psi}{\delta}$ plots in Fig. \ref{variationsAlpha:error}. It can be seen in this figure that at the end of the engagement, both these quantities lie within the bounds $([-1, 1])$ which are shown by dashed blue lines.

\begin{figure}[!htbp]
    \centering
\subfigure[Trajectories for different $v_T$]{\includegraphics[width=0.32\textwidth]{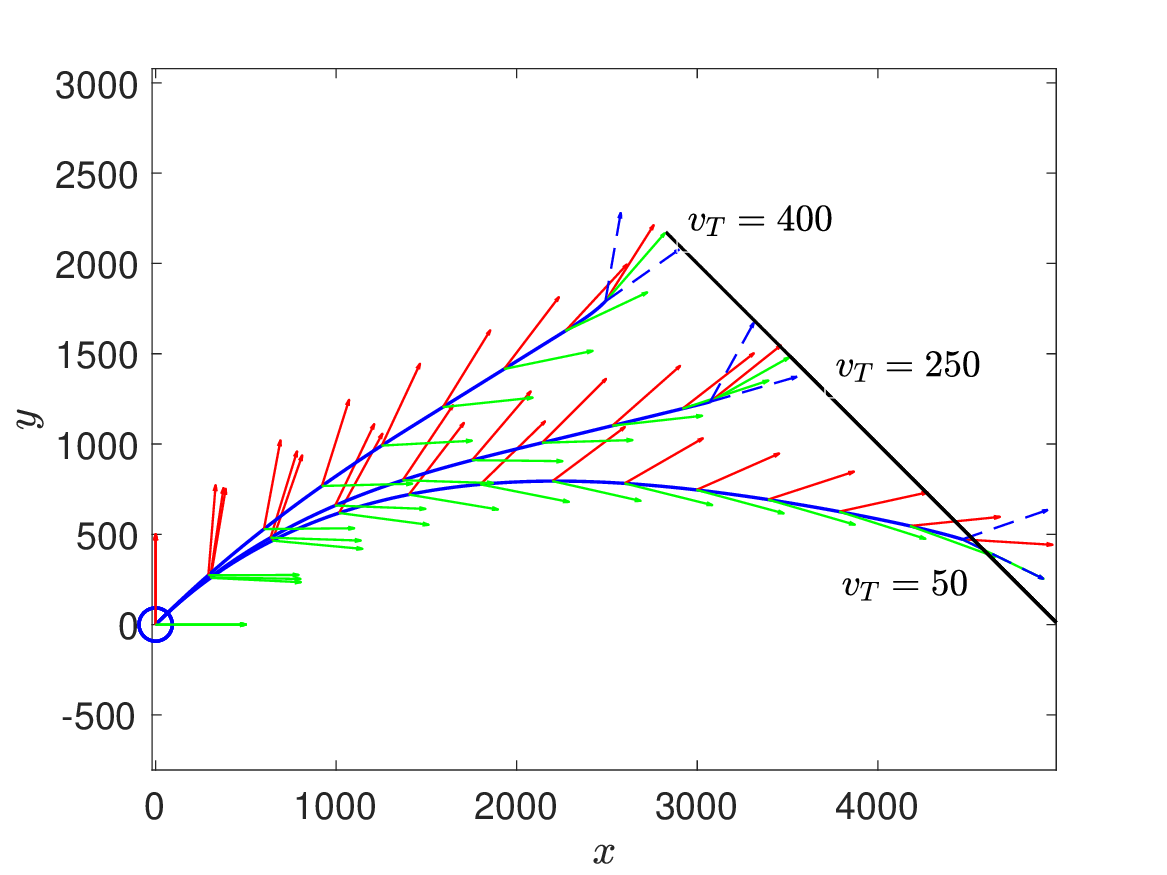}\label{variationsSpeed:Trajectory}}
\subfigure[Trajectories for different $a_T$]{\includegraphics[width=0.32\textwidth]{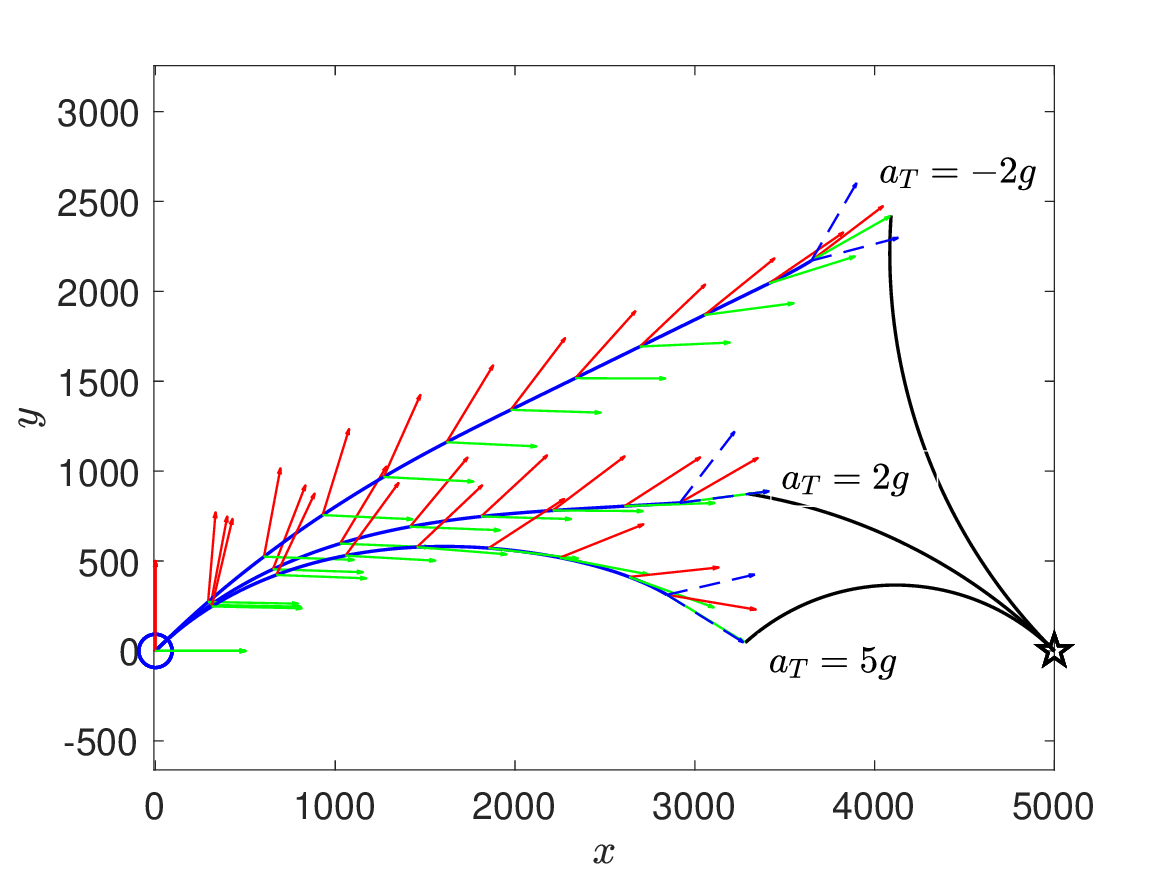}\label{variationsAccel:Trajectory}}
\subfigure[Trajectories for different $\theta_{T0}$]{\includegraphics[width=0.32\textwidth]{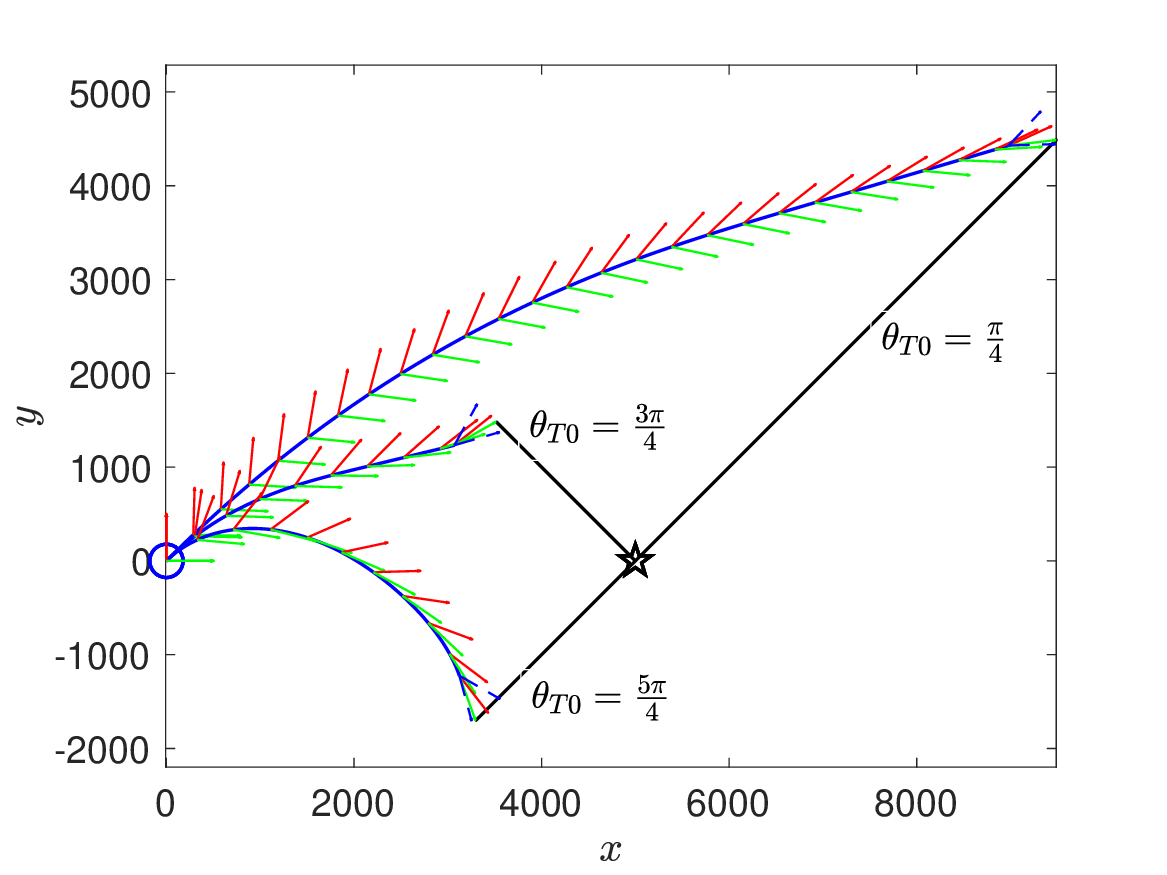}\label{variationsAngle:Trajectory}}
\subfigure[Predicted intercept time for different $v_T$]{\includegraphics[width=0.32\textwidth]{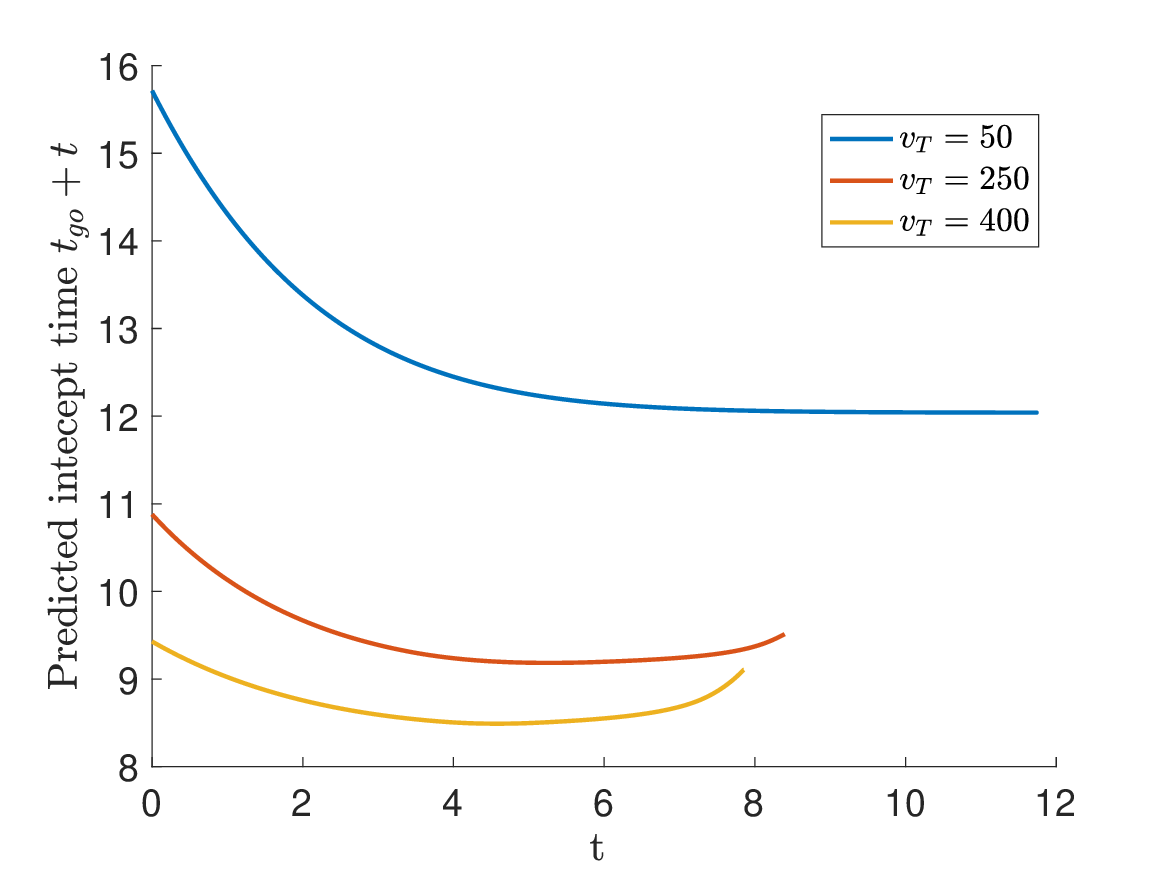}\label{variationsSpeed:Tgo}}
\subfigure[Predicted intercept time for different  $a_T$]{\includegraphics[width=0.32\textwidth]{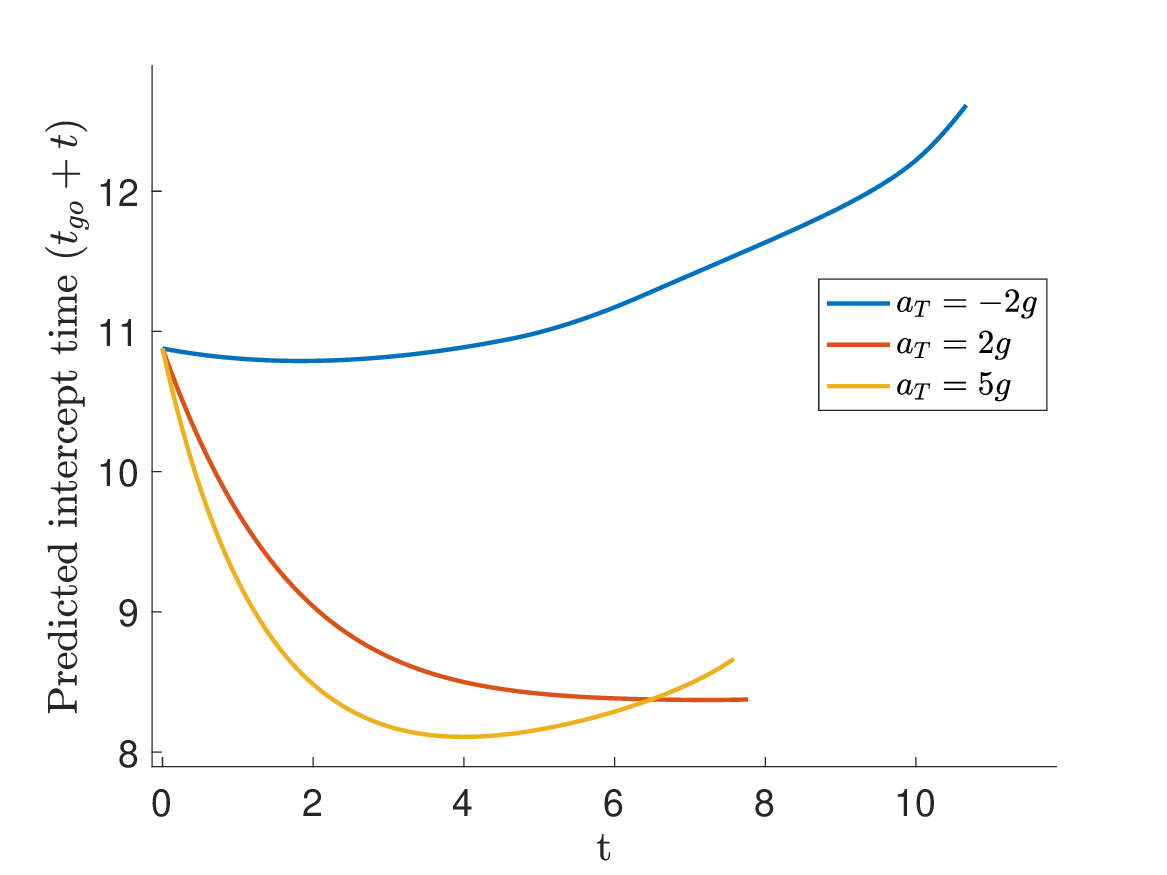}\label{variationsAccel:Tgo}}
\subfigure[Predicted intercept time for different $\theta_{T0}$]{\includegraphics[width=0.32\textwidth]{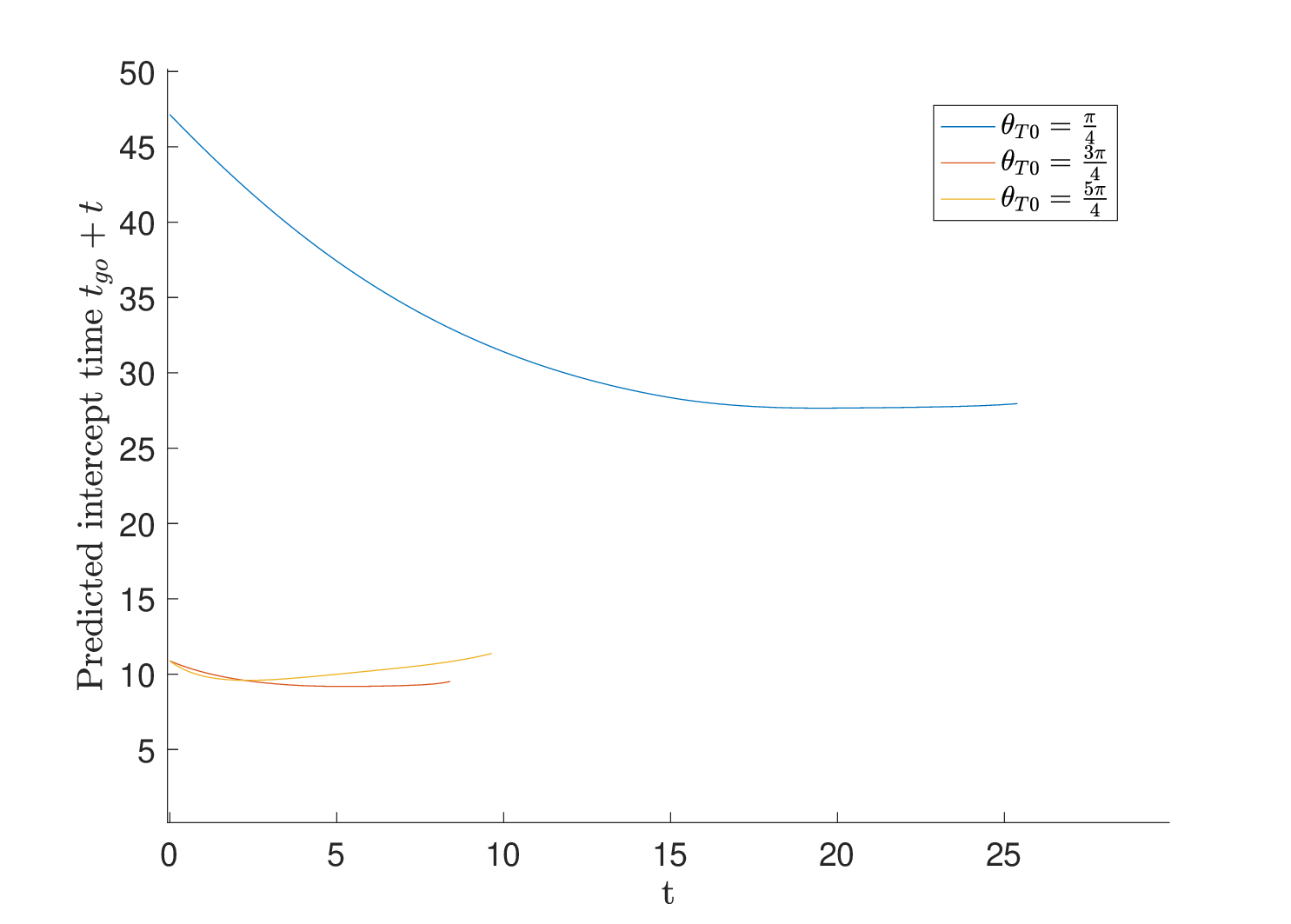}\label{variationsAngle:Tgo}}
    \caption{Variation of Trajectories with Target's Speed, Acceleration, and Launch Angles}
    \label{variations}
\end{figure}

Fig. \ref{variations} illustrates successful capture by the pursuer-turret assembly for different speeds (Fig. \ref{variationsSpeed:Trajectory}), accelerations (Fig. \ref{variationsAccel:Trajectory}), and launch angles of the target (Fig. \ref{variationsAngle:Trajectory}). Note that, even if the pursuer-turret assembly and the target are not initially on the collision course, due to the analytic expression of the guidance law, the non-linear engagement kinematics gets iteratively linearized at each time instance when the guidance command is computed. Figs. \ref{variationsSpeed:Tgo}-\ref{variationsAngle:Tgo} show the corresponding change in the predicted intercept time $(t_{go}+t)$ throughout the engagement. The approximation of $t_{go}$ in Eq. \eqref{tgoApprox} is exact only if a non-accelerating pursuer and a non-accelerating target are on the collision course. In this case the linearization holds perfectly. But when this is not true, one can notice from Figs. \ref{variationsSpeed:Tgo}-\ref{variationsAngle:Tgo} that initially the prediction of intercept time is not accurate. But towards the end of the engagement, when the puruser maneuvers to be on the collision course the predicted intercept time does not vary significantly. This shows that the guidance law steers the pursuer towards near collision course conditions.  This leads to successful capture in cases when the pursuer-turret assembly is not initially on the collision course to the target.

\newpage
\section{Conclusion}
A linear quadratic guidance law for joint motion planning of a vehicle with an attached rotating turret is presented. The turret has a limited range as well as a field of view. The objective is to capture a maneuvering target such that at the terminal time, it is within the field-of-view and range limits. By minimizing the convex sum of the control efforts of the pursuer and the turret, the guidance law provides an ability to compensate for the turn limitations of the pursuer and the turret by tuning the parameter $\alpha$. The guidance law is obtained using linearization about the collision triangle and admits an analytical solution that facilitates its onboard implementation. Simulation results are presented to exemplify the cooperation between the turret and the vehicle and show the performance of the guidance law for different engagement scenarios.
The results demonstrate the efficacy of the guidance law even when the initial configuration of the engagement varies significantly from the collision triangle.
As the approximation of time-to-go used in this work is exact only for the case when the pursuer and target pair are on the collision course, a better time-to-go approximation specific to the proposed guidance is expected to improve the performance. In some scenarios, it is imperative that the actuators of the  pursuer-turret assembly satisfy hard constraints. These directions can be considered for future works.

\section*{Funding Sources}

This research was supported by the Aerospace Systems Technology Research and Assessment (ASTRA) Aerospace Technology Development and Testing (ATDT) program at AFRL under contract number FA865021D2602.

 \section*{Acknowledgments}
 The authors would like to thank Dr. Isaac Weintraub, Dr. Satyanarayana Gupta Manyam, and Dr. Shivam Bajaj for fruitful discussions.  

\bibliography{sample}

\begin{thebibliography}{29}
\newcommand{\enquote}[1]{``#1''}
\providecommand{\natexlab}[1]{#1}
\providecommand{\url}[1]{\texttt{#1}}
\providecommand{\urlprefix}{URL }
\expandafter\ifx\csname urlstyle\endcsname\relax
  \providecommand{\doi}[1]{\discretionary{}{}{}https://doi.org/#1}\else
  \providecommand{\doi}[1]{\discretionary{}{}{}\urlstyle{rm}\url{https://doi.org/#1}}\fi

\bibitem[{Schonberger et~al.(1982)Schonberger, Fuhs, and
  Mandigo}]{schonberger1982flow}
Schonberger, J.~R., Fuhs, A.~E., and Mandigo, A.~M., \enquote{Flow control for
  an airborne laser turret,} \emph{Journal of Aircraft}, Vol.~19, No.~7, 1982,
  pp. 531--537.
\newblock \doi{10.2514/6.1981-1637}.

\bibitem[{Craig and CA(1981)}]{craig1981study}
Craig, J., and CA, S. D. L. I. C.~M., \enquote{A Study in Flow Control and
  Screening Methods for Aircraft Laser Turrets,} \emph{Air Force Report}, 1981.

\bibitem[{Gupta(1999)}]{gupta1999evaluation}
Gupta, A., \enquote{Evaluation of a fully assembled armored vehicle
  hull--turret model using computational and experimental modal analyses,}
  \emph{Computers \& structures}, Vol.~72, No. 1-3, 1999, pp. 177--183.
\newblock \doi{https://doi.org/10.1016/s0045-7949(99)00024-3}.

\bibitem[{dos Santos~Gomes and Ferreira(2005)}]{dos2005gun}
dos Santos~Gomes, M., and Ferreira, A.~M., \enquote{Gun-turret modelling and
  control,} \emph{ABCM Symposium Series in Mechatronics}, Vol.~2, 2005, pp.
  60--67.

\bibitem[{Bryson~Jr.(1965)}]{bryson1965linear}
Bryson~Jr., A.~E., \enquote{Linear feedback solutions for minimum effort
  interception, rendezvous, and soft landing,} \emph{AIAA Journal}, Vol.~3,
  No.~8, 1965, pp. 1542--1544.
\newblock \doi{10.2514/3.3199}.

\bibitem[{Kreindler(1973)}]{kreindler1973optimality}
Kreindler, E., \enquote{Optimality of proportional navigation,} \emph{AIAA
  Journal}, Vol.~11, No.~6, 1973, pp. 878--880.
\newblock \doi{10.2514/3.50527}.

\bibitem[{Nesline and Zarchan(1981)}]{nesline1981new}
Nesline, F.~W., and Zarchan, P., \enquote{A new look at classical vs modern
  homing missile guidance,} \emph{Journal of Guidance and Control}, Vol.~4,
  No.~1, 1981, pp. 78--85.
\newblock \doi{10.2514/3.56054}.

\bibitem[{Shaferman and Shima(2008)}]{shaferman2008linear}
Shaferman, V., and Shima, T., \enquote{Linear quadratic guidance laws for
  imposing a terminal intercept angle,} \emph{Journal of Guidance, Control, and
  Dynamics}, Vol.~31, No.~5, 2008, pp. 1400--1412.
\newblock \doi{10.2514/1.32836}.

\bibitem[{Gutman(1979)}]{gutman1979optimal}
Gutman, S., \enquote{On optimal guidance for homing missiles,} \emph{Journal of
  Guidance and Control}, Vol.~2, No.~4, 1979, pp. 296--300.
\newblock \doi{10.2514/3.55878}.

\bibitem[{Weiss and Shima(2018)}]{weiss2018linear}
Weiss, M., and Shima, T., \enquote{Linear quadratic optimal control-based
  missile guidance law with obstacle avoidance,} \emph{IEEE Transactions on
  Aerospace and Electronic Systems}, Vol.~55, No.~1, 2018, pp. 205--214.
\newblock \doi{10.1109/TAES.2018.2849901}.

\bibitem[{Tsukerman et~al.(2018)Tsukerman, Weiss, Shima, L{\"o}bl, and
  Holzapfel}]{tsukerman2018optimal}
Tsukerman, A., Weiss, M., Shima, T., L{\"o}bl, D., and Holzapfel, F.,
  \enquote{Optimal rendezvous guidance laws with application to civil
  autonomous aerial refueling,} \emph{Journal of guidance, control, and
  dynamics}, Vol.~41, No.~5, 2018, pp. 1167--1174.
\newblock \doi{10.2514/1.G003154}.

\bibitem[{Tan et~al.(2018)Tan, Fonod, and Shima}]{tan2018cooperative}
Tan, Z.~W., Fonod, R., and Shima, T., \enquote{Cooperative guidance law for
  target pair to lure two pursuers into collision,} \emph{Journal of Guidance,
  Control, and Dynamics}, Vol.~41, No.~8, 2018, pp. 1687--1699.
\newblock \doi{10.2514/1.G003357}.

\bibitem[{Jha et~al.(2019)Jha, Tsalik, Weiss, and Shima}]{jha2019cooperative}
Jha, B., Tsalik, R., Weiss, M., and Shima, T., \enquote{Cooperative guidance
  and collision avoidance for multiple pursuers,} \emph{Journal of Guidance,
  Control, and Dynamics}, Vol.~42, No.~7, 2019, pp. 1506--1518.
\newblock \doi{10.2514/1.G004139}.

\bibitem[{Obermeyer(2009)}]{obermeyer2009path}
Obermeyer, K., \enquote{Path planning for a UAV performing reconnaissance of
  static ground targets in terrain,} \emph{AIAA guidance, navigation, and
  control conference}, 2009, p. 5888.
\newblock \doi{10.2514/6.2009-5888}.

\bibitem[{Dumitrescu and Mitchell(2003)}]{dumitrescu2003approximation}
Dumitrescu, A., and Mitchell, J.~S., \enquote{Approximation algorithms for TSP
  with neighborhoods in the plane,} \emph{Journal of Algorithms}, Vol.~48,
  No.~1, 2003, pp. 135--159.
\newblock \doi{10.1016/S0196-6774(03)00047-6}.

\bibitem[{Obermeyer et~al.(2010)Obermeyer, Oberlin, and
  Darbha}]{obermeyer2010sampling}
Obermeyer, K., Oberlin, P., and Darbha, S., \enquote{Sampling-based roadmap
  methods for a visual reconnaissance UAV,} \emph{AIAA guidance, navigation,
  and control conference}, 2010, p. 7568.
\newblock \doi{10.2514/6.2010-7568}.

\bibitem[{Isaacs et~al.(2011)Isaacs, Klein, and
  Hespanha}]{isaacs2011algorithms}
Isaacs, J.~T., Klein, D.~J., and Hespanha, J.~P., \enquote{Algorithms for the
  traveling salesman problem with neighborhoods involving a Dubins vehicle,}
  \emph{Proceedings of the 2011 American control conference}, IEEE, 2011, pp.
  1704--1709.
\newblock \doi{10.1109/ACC.2011.5991501}.

\bibitem[{Weiss and Shima(2016)}]{weiss2016minimum}
Weiss, M., and Shima, T., \enquote{Minimum effort pursuit/evasion guidance with
  specified miss distance,} \emph{Journal of Guidance, Control, and Dynamics},
  Vol.~39, No.~5, 2016, pp. 1069--1079.
\newblock \doi{10.2514/1.G001623}.

\bibitem[{Dubins(1957)}]{dubins1957curves}
Dubins, L.~E., \enquote{On curves of minimal length with a constraint on
  average curvature, and with prescribed initial and terminal positions and
  tangents,} \emph{American Journal of mathematics}, Vol.~79, No.~3, 1957, pp.
  497--516.
\newblock \doi{10.2307/2372560}.

\bibitem[{Ratnoo and Ghose(2008)}]{ratnoo2008impact}
Ratnoo, A., and Ghose, D., \enquote{Impact angle constrained interception of
  stationary targets,} \emph{Journal of Guidance, Control, and Dynamics},
  Vol.~31, No.~6, 2008, pp. 1817--1822.
\newblock \doi{10.2514/1.37864}.

\bibitem[{Gopalan et~al.(2017)Gopalan, Ratnoo, and
  Ghose}]{gopalan2017generalized}
Gopalan, A., Ratnoo, A., and Ghose, D., \enquote{Generalized time-optimal
  impact-angle-constrained interception of moving targets,} \emph{Journal of
  Guidance, Control, and Dynamics}, Vol.~40, No.~8, 2017, pp. 2115--2120.
\newblock \doi{10.2514/1.G002384}.

\bibitem[{Ratnoo and Ghose(2009)}]{ratnoo2009state}
Ratnoo, A., and Ghose, D., \enquote{State-dependent Riccati-equation-based
  guidance law for impact-angle-constrained trajectories,} \emph{Journal of
  Guidance, Control, and Dynamics}, Vol.~32, No.~1, 2009, pp. 320--326.
\newblock \doi{10.2514/1.37876}.

\bibitem[{Indig et~al.(2016)Indig, Ben-Asher, and Sigal}]{indig2016near}
Indig, N., Ben-Asher, J.~Z., and Sigal, E., \enquote{Near-optimal minimum-time
  guidance under spatial angular constraint in atmospheric flight,}
  \emph{Journal of Guidance, Control, and Dynamics}, Vol.~39, No.~7, 2016, pp.
  1563--1577.
\newblock \doi{10.2514/6.2015-0089}.

\bibitem[{V{\'a}na and Faigl(2018)}]{vana2018optimal}
V{\'a}na, P., and Faigl, J., \enquote{Optimal Solution of the Generalized
  Dubins Interval Problem,} \emph{Robotics: Science and Systems}, 2018.

\bibitem[{Manyam et~al.(2017)Manyam, Rathinam, Casbeer, and
  Garcia}]{manyam2017tightly}
Manyam, S.~G., Rathinam, S., Casbeer, D., and Garcia, E., \enquote{Tightly
  bounding the shortest Dubins paths through a sequence of points,}
  \emph{Journal of Intelligent \& Robotic Systems}, Vol.~88, 2017, pp.
  495--511.
\newblock \doi{10.1007/s10846-016-0459-4}.

\bibitem[{Garber(1968)}]{garber1968optimum}
Garber, V., \enquote{Optimum intercept laws for accelerating targets.}
  \emph{AIAA Journal}, Vol.~6, No.~11, 1968, pp. 2196--2198.
\newblock \doi{10.2514/3.4962}.

\bibitem[{Rubinsky and Gutman(2014)}]{rubinsky2014three}
Rubinsky, S., and Gutman, S., \enquote{Three-player pursuit and evasion
  conflict,} \emph{Journal of Guidance, Control, and Dynamics}, Vol.~37, No.~1,
  2014, pp. 98--110.
\newblock \doi{10.2514/1.61832}.

\bibitem[{Bryson and Ho(2018)}]{bryson2018applied}
Bryson, A.~E., and Ho, Y.-C., \emph{Applied optimal control: optimization,
  estimation, and control}, Routledge, 2018, Chap.~5.
\newblock \doi{10.1201/9781315137667}.

\bibitem[{Pontryagin(1987)}]{pontryagin1987mathematical}
Pontryagin, L.~S., \emph{Mathematical theory of optimal processes}, CRC press,
  1987.
\newblock \doi{10.1201/9780203749319}.

\end{thebibliography}

\end{document}